%% file: main.tex
\documentclass[conference]{IEEEtran}

\usepackage{cite}
\usepackage{amsmath,amssymb,amsfonts}
\usepackage{booktabs,tabularx,tabulary,array}
\newcolumntype{Y}{>{\centering\arraybackslash}X}
\newcolumntype{L}{>{\raggedright\arraybackslash}p{0.26\textwidth}}
\usepackage{graphicx}
\usepackage[caption=false,font=footnotesize]{subfig}
\usepackage{comment}
\usepackage{tablefootnote}
\usepackage{circuitikz}
\usepackage{xspace}
\usepackage{siunitx}
\usepackage{multirow}
\usepackage{marvosym}
\usepackage{diagbox}
\usepackage{makecell}
\usepackage[table]{xcolor}
\usepackage{threeparttable}
\usepackage{enumitem}
\usepackage{float}
\usepackage[official]{eurosym}
\usepackage[english]{babel}
\usepackage{svg}
\usepackage[version=4]{mhchem}
\usepackage{tikz}
\usetikzlibrary{arrows.meta, positioning}
\usepackage[hidelinks]{hyperref}

\definecolor{LightRed}{rgb}{1, 0.6, 0.6}
\definecolor{LightGreen}{rgb}{0.0, 0.8, 0.6}
\definecolor{LightBlue}{rgb}{0.61, 0.87, 1.0}
\definecolor{Grey}{rgb}{0.66, 0.66, 0.66}

\def\BibTeX{{\rm B\kern-.05em{\sc i\kern-.025em b}\kern-.08em
    T\kern-.1667em\lower.7ex\hbox{E}\kern-.125emX}}

\renewcommand\IEEEkeywordsname{Keywords}

\begin{document}

\title{Animal Re-Identification on Microcontrollers}

\author{\IEEEauthorblockN{Yubo Chen, Di Zhao, Yun Sing Koh, and Talia Xu}
\IEEEauthorblockA{\textit{University of Auckland}\\
Auckland, New Zealand\\
ych259@aucklanduni.ac.nz, di.zhao@auckland.ac.nz, y.koh@auckland.ac.nz, talia.xu@auckland.ac.nz}
}

\maketitle

\begin{abstract}
Camera-based Animal Re-identification (Animal Re-ID) can support wildlife monitoring and precision livestock management where wireless connectivity is limited. In such deployments, inference should run on collar tags or low-power edge nodes built around microcontrollers (MCUs), yet most Animal Re-ID models are too large and assume higher-resolution inputs.
We study how to make Animal Re-ID practical on MCU-class hardware through a deployment-driven investigation of model compression under low-resolution sensing and strict memory constraints. First, we show that feature-level knowledge distillation from large teachers provides limited benefit once memory, architecture mismatch, and input resolution are considered together. Guided by this result, we design a compact CNN-based Re-ID model by scaling the width and depth of a MobileNetV2 backbone for low-resolution inputs. We then evaluate the model across six public Animal Re-ID datasets and a self-collected cattle dataset, and introduce a data-efficient fine-tuning strategy that adapts to a new site with three images per identity.
The final model reduces size by over two orders of magnitude while preserving competitive retrieval accuracy. On Arduino deployment, it runs fully on-device with only a small mAP drop and unchanged Top-1 accuracy relative to the cluster version.
\end{abstract}


\input{sections/introduction}

\input{sections/background}
\input{sections/initial_study}
\input{sections/pruning}

\input{sections/experiment_on_arduino}

\input{sections/discussion}

\input{sections/related_work}
\input{sections/conclusion}

\bibliographystyle{IEEEtran}
\bibliography{references}

\end{document}

%% file: sections/introduction.tex
\vspace{-3mm}
\section{Introduction}
\vspace{-1mm}

Tracking individual animals is important for wildlife monitoring, conservation, and precision livestock management, where repeated sightings must be linked to the same identity over time~\cite{kabuga2024similarity, guo2025individual}. Animal Re-identification (Animal Re-ID) automates this process by matching a newly captured query image against a gallery of known individuals. However, accurate Animal Re-ID is challenging because different individuals can appear visually similar, while the same individual may be observed under changing pose, lighting, background, and camera conditions.

Recent high-performing Re-ID systems increasingly rely on large vision transformers and vision--language models~\cite{li2023clip}. These models are accurate but expensive: their memory and compute requirements usually confine inference to servers or workstations~\cite{saha2025vision, ahmed2025deepcompress, zhong2023lightweight}. This creates a mismatch with field deployments, where images are captured by camera traps, collars, or other edge devices but must often be transferred elsewhere for analysis. In remote areas, unreliable connectivity and the cost of transmitting raw images can delay or prevent timely decisions~\cite{velasco2024reliable, scherer2022widevision, kaltenbach2025can}. This motivates moving Re-ID inference closer to the sensor, reducing communication overhead and enabling faster local identification.

We target Animal Re-ID directly on microcontroller-class edge devices. This setting introduces two coupled challenges. First, standard MCUs provide only hundreds of kilobytes to a few megabytes of memory, far below the footprint of many state-of-the-art Animal Re-ID models~\cite{heydari2025tiny, ray2022review}. Second, low-cost embedded cameras often produce low-resolution, noisy images that lack the fine-grained texture available in curated datasets~\cite{scherer2022widevision, mulero2025addressing}. The core problem is therefore to design a model that is small enough for MCU deployment while still producing discriminative embeddings from degraded animal imagery.

In this work, we study how to make Animal Re-ID feasible under MCU and low-resolution sensing constraints. The novelty is not a new compression algorithm in isolation, but a deployment-driven study of Animal Re-ID in a rarely explored setting combining MCU-class hardware, low-resolution sensing, strict Flash/SRAM budgets, and on-device retrieval. Our contributions are threefold. \textit{\textbf{First}}, we show that feature-level knowledge distillation provides limited benefit when architecture mismatch, input resolution, and memory limits are considered together (Section~\ref{sec:sec3-initial-study}). \textit{\textbf{Second}}, we systematically scale the width and depth of a MobileNetV2-based Re-ID model and identify a compact configuration for $64\times64$ inputs (Section~\ref{sec:sec4-pruning}). \textit{\textbf{Third}}, we deploy the quantised model on an Arduino Nano 33 BLE Sense and evaluate the complete on-device Re-ID pipeline on a real cattle dataset, including few-shot adaptation to a new site (Section~\ref{sec:sec5-experiment-arduino}).


%% file: sections/background.tex
\section{System Constraints and Setup}
\label{sec:background}

\subsection{Hardware Constraints}

We target the Arduino Nano 33 BLE Sense as a representative MCU-class platform for low-power Animal Re-ID. The device has 1\,MB of Flash, of which about 960\,KB is usable by the application, and 256\,KB of SRAM. These limits make both server-scale Re-ID models and high-resolution image processing infeasible: a single $1920\times1080\times3$ 8-bit image would require more than 6\,MB of memory, far exceeding the available SRAM.

\begin{figure}[t]
    \centering
    \vspace{-4mm}
    \subfloat[High resolution\label{fig:arduino_sample}]{%
        \includegraphics[height=2.6cm]{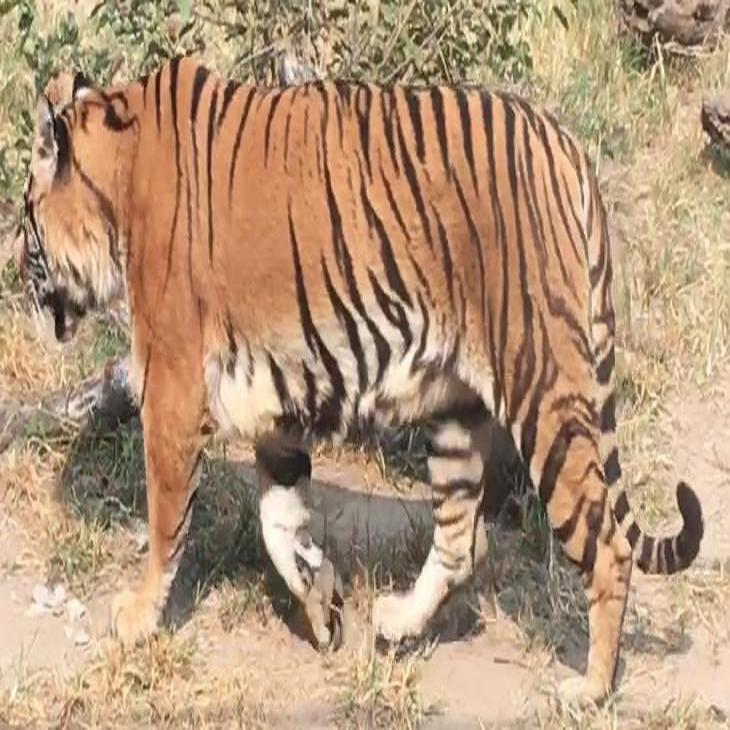}}%
    \hspace{5mm}
    \subfloat[Low resolution\label{fig:friesian_sample}]{%
        \includegraphics[height=2.6cm]{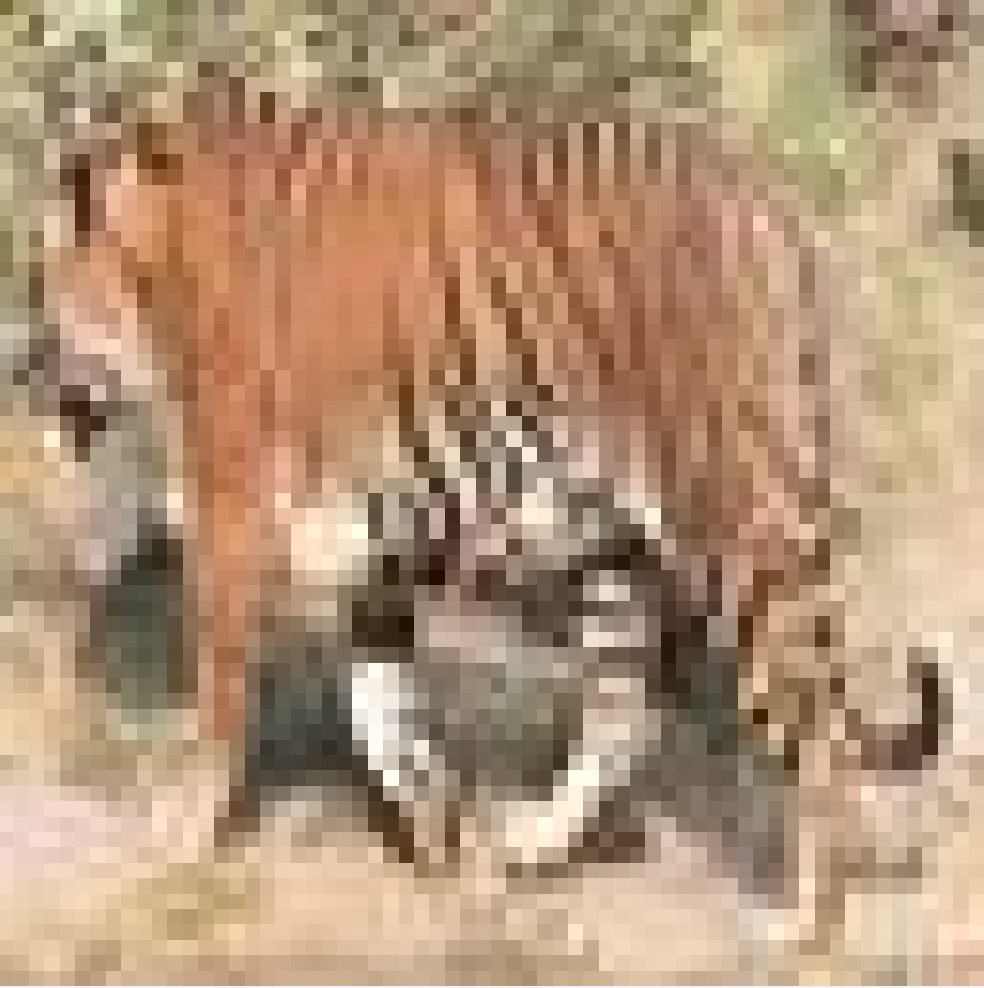}}
    \caption{High- and low-resolution examples from ATRW.}
    \vspace{-6mm}
    \label{fig:image_quality_comparison}
\end{figure}

To align the model with this deployment setting, all images are resized to $64\times64\times3$, requiring roughly 12\,KB per frame. This resolution is also consistent with low-cost embedded camera modules such as the OV7675, whose outputs are affected by noise, colour distortion, and limited dynamic range, as illustrated in \autoref{fig:image_quality_comparison}. On the SRAM side, our deployment uses a 120\,KB TensorFlow Lite Micro tensor arena for activations, temporary buffers, and runtime metadata. On the Flash side, after reserving space for program code and gallery embeddings, only about 600--700\,KB can be allocated to the deployed Re-ID model. These constraints determine the input resolution, model footprint, and compression strategy used in the rest of the paper.

\vspace{-2mm}
\subsection{Animal Datasets}

We evaluate the proposed framework on both public Animal Re-ID benchmarks and a self-collected deployment dataset. The public benchmark set contains six datasets: ATRW, FriesianCattle2017, LionData, MPDD, IPanda50, and CoBRA ReID Youngstock~\cite{li2019atrw,he2023animal,andrew2017visual,dlamini2020automated,wang2021giant,perneel2025dynamic}. Together, these datasets cover multiple species, including tigers, pandas, lions, cattle, and dogs, and provide diverse imaging conditions for evaluating cross-species generalisation.

To test real-world deployment, we also collect a local cattle Re-ID dataset in Auckland, New Zealand, which is described in detail in Section~\ref{sec:sec5-experiment-arduino}. This dataset introduces a domain shift from public benchmarks due to differences in breed, lighting, background, viewpoint, and capture conditions. We use it to evaluate whether the model can be adapted to a new site with only a small number of local reference images, rather than requiring full retraining.

All datasets follow the standard Re-ID training--gallery--query protocol. The training set is used offline to learn the embedding model. The gallery set represents registered animals stored on the device as reference embeddings. The query set represents new sightings, which are matched against the gallery using nearest-neighbour retrieval in the learned embedding space.

%% file: sections/initial_study.tex
\vspace{-3mm}
\section{Initial Study}
\label{sec:sec3-initial-study}

A natural approach for fitting Animal Re-ID onto MCU-class devices is to start from a stronger server-side model and compress it through knowledge distillation (KD), where a compact student learns from a larger teacher~\cite{hinton2015distilling,moslemi2024survey}. This is appealing because recent Re-ID models can produce strong identity embeddings, while their memory footprint prevents direct MCU deployment. We therefore first test whether KD can provide a practical path from accurate server-scale models to MCU-ready students.

We compare two teachers, CLIP-ReID~\cite{li2023clip} and Few-Shot Animal Re-ID~\cite{wahltinez2024open}, with two compact students, MobileNetV2~\cite{sandler2018mobilenetv2} and ViT-Tiny~\cite{touvron2021training}. CLIP-ReID represents a Transformer-based teacher, Few-Shot Animal Re-ID provides a CNN-based teacher, MobileNetV2 is our main MCU-friendly CNN student, and ViT-Tiny tests whether a smaller Transformer can better match the CLIP-ReID teacher. This setup lets us examine whether KD is limited by teacher strength, teacher--student architecture mismatch, model size, or low-resolution inputs.

\subsection{Knowledge Distillation}

In our KD experiments, we study three variants: ViT$\rightarrow$CNN, ViT$\rightarrow$ViT, and CNN$\rightarrow$CNN. The datasets follow the training--gallery--query split described in \autoref{sec:background}. Each model is evaluated using mean Average Precision (mAP) and Rank-1, Rank-5, and Rank-10 retrieval accuracy. mAP measures how well the model retrieves all correct gallery matches for each query~\cite{zheng2015scalable}.

\begin{table*}[t]
\centering
\footnotesize
\renewcommand{\arraystretch}{1.08}
\setlength{\tabcolsep}{4pt}
\caption{Summary of distillation outcomes. Representative mAP values are from ATRW, FriesianCattle2017, and LionData.}
\vspace{-2mm}
\label{tab:kd-summary}
\begin{tabularx}{\textwidth}{>{\raggedright\arraybackslash}p{0.20\textwidth}
>{\raggedright\arraybackslash}p{0.07\textwidth}
>{\raggedright\arraybackslash}p{0.31\textwidth}
>{\raggedright\arraybackslash}X}
\toprule
Route & Student & Representative result & Implication \\
\midrule
CLIP-ReID $\rightarrow$ MobileNetV2 & CNN & Direct MobileNetV2 beats the distilled student (51.2/95.3/35.7 vs.\ 40.6/85.0/23.4). & Cross-architecture feature matching hurts the MCU-friendly CNN baseline. \\
CLIP-ReID $\rightarrow$ ViT-Tiny & ViT & ViTKD/TeKAP preserve much of the teacher performance (ATRW mAP 53.8/55.4 vs.\ 58.2). & Architecture-aligned KD works, but the ViT student is about 22 MB and remains too large. \\
Few-Shot $\rightarrow$ MobileNetV2 & CNN & Direct MobileNetV2 again outperforms the distilled student (51.2/95.3/35.7 vs.\ 50.0/90.9/21.1). & The teacher is not consistently stronger than the lightweight student under our retrieval protocol. \\
\bottomrule
\end{tabularx}
\vspace{-2mm}
\end{table*}

The results in \autoref{tab:kd-summary} show that feature-level KD is not sufficient for MCU-ready Animal Re-ID. In the ViT$\rightarrow$CNN setting, distillation from CLIP-ReID to MobileNetV2 does not improve the student, suggesting that forcing a compact CNN to reproduce Transformer features can hurt the MCU-friendly baseline. In the ViT$\rightarrow$ViT setting, distillation is more effective because the teacher and student are architecturally aligned, but the resulting ViT-Tiny student is still about 22\,MB and remains far beyond the target memory budget. In the CNN$\rightarrow$CNN setting, distillation from the Few-Shot teacher again fails to improve over directly trained MobileNetV2, indicating that the teacher is not consistently stronger under our retrieval protocol.

Overall, KD helps only in the case where the student remains too large for deployment, while the deployable CNN student is better trained directly. We therefore shift from teacher-driven compression to direct structural design: starting from MobileNetV2, we study how much width and depth can be removed while preserving retrieval accuracy under low-resolution inputs.

\subsection{Low-Resolution Input}

The KD experiments above focus on model compression under relatively favourable imaging conditions, where inputs are resized to $224 \times 224 \times 3$. In realistic MCU deployments, however, the input itself is also constrained. Limited SRAM restricts intermediate activation memory, and low-cost embedded cameras often produce small, noisy images rather than high-quality inputs. A model that performs well at standard resolution may therefore lose retrieval accuracy when deployed with MCU-scale sensing.

To isolate this effect, we evaluate representative teacher models under both high-resolution and low-resolution inputs. Specifically, we resize the input images to $64 \times 64 \times 3$, matching the deployment setting described in \autoref{sec:background}, and compare this with the original $224 \times 224 \times 3$ setting. We keep the same training and evaluation protocol and report results on three datasets using CLIP-ReID and Few-Shot Animal Re-ID.

\begin{table*}
\vspace{-2mm}
\centering
\footnotesize
\renewcommand{\arraystretch}{1.1}
\setlength{\tabcolsep}{2pt}
\caption{Teacher performance under high- and low-resolution inputs. Shaded rows indicate low-resolution inputs.}
\vspace{-2mm}
\label{tab:teacher_high_low}
\begin{tabularx}{\textwidth}{lYYYYYYYYYYYY}
\toprule
& \multicolumn{4}{c}{ATRW} & \multicolumn{4}{c}{FriesianCattle2017} & \multicolumn{4}{c}{LionData} \\
\cmidrule(lr){2-5} \cmidrule(lr){6-9} \cmidrule(lr){10-13}
Method & mAP & Top-1 & Top-5 & Top-10 & mAP & Top-1 & Top-5 & Top-10 & mAP & Top-1 & Top-5 & Top-10 \\
\midrule
CLIP-ReID (high-res)      & 58.2 & 94.6 & 98.6 & 99.3 & 89.0 & 97.6 & 100.0 & 100.0 & 23.9 & 29.5 & 68.9 & 86.9 \\
\rowcolor{blue!10}
CLIP-ReID (low-res)       & 50.3 & 92.5 & 98.8 & 99.3 & 73.4 & 88.2 & 100.0 & 100.0 & 17.7 & 21.3 & 50.8 & 75.4 \\
Few-Shot (high-res)       & 45.5 & 87.5 & 96.5 & 98.1 & 84.6 & 91.8 & 100.0 & 100.0 & 20.3 & 21.3 & 53.7 & 72.1 \\
\rowcolor{blue!10}
Few-Shot (low-res)        & 33.6 & 73.1 & 90.6 & 94.1 & 59.1 & 80.0 & 87.1 & 92.9 & 19.3 & 23.0 & 50.8 & 68.9 \\
\bottomrule
\end{tabularx}
\vspace{-5mm}
\end{table*}

The results are reported in~\autoref{tab:teacher_high_low}. Compared with their high-resolution counterparts, both models generally obtain lower mAP and Top-1/Top-$K$ accuracy under low-resolution inputs. The drop is especially clear on FriesianCattle2017, where CLIP-ReID mAP decreases from 89.0 to 73.4 and Few-Shot mAP decreases from 84.6 to 59.1. This shows that reducing image resolution weakens fine-grained identity discrimination and makes gallery ranking less reliable, consistent with prior work on low-resolution Re-ID~\cite{zhang2021low,jiao2018deep,jing2015super}.

Together, these diagnostic studies motivate the architecture design in the next section. KD does not yield a student that is both accurate and small enough for MCU deployment, and low-resolution inputs reduce the effectiveness of models designed for standard image sizes. We therefore directly customise a compact CNN backbone for $64\times64$ Animal Re-ID under MCU memory constraints.

%% file: sections/pruning.tex
\section{Structurally Guided Width--Depth Pruning for Customizing MobileNetV2}
\label{sec:sec4-pruning}

Our initial experiments in \autoref{sec:sec3-initial-study} show that, in a constrained hardware setting, knowledge distillation offers only diminishing returns. We also find that reducing the input resolution to $64\times64$ leads to a significant drop in accuracy across all tested architectures.
This degradation indicates that the fine-grained spatial details that standard models are designed to exploit are largely absent in low-resolution inputs. 
Among the lightweight models, MobileNetV2 stands out as a deployment-friendly baseline. It achieves the best accuracy–complexity trade-off, uses operators supported by common MCU ML runtimes (TensorFlow Lite Micro), however, the standard implementation still exceeds the memory budget of our target MCU and is tuned for high-resolution inputs. 
Motivated by these observations, we shift our focus from a teacher-based distillation to direct structural optimisation. By systematically scaling the MobileNetV2’s structural dimensions, we aim to better match the information content of low-resolution inputs while meeting the stringent memory constraints of the MCU.


\subsection{Baseline MobileNetV2 Architecture}
    
We first summarise the features of the baseline MobileNetV2 ~\cite{sandler2018mobilenetv2} architecture that are important for our structural depth and width scaling. Our implementation follows the Keras reference configuration~\cite{keras_mobilenet_api}, with minor changes to accept $64\times64$ inputs instead of the standard $224\times224$. 
A $64\times64\times3$ input image is first passed through a $3\times3$ convolutional stem with a stride of 2, which increases the channel count and reduces the resolution to $32\times32$. For convenience, we refer to the number of channels in a feature map as the network width. The rest of the backbone is a simple chain of inverted residual bottleneck blocks. Each block: (1) starts from a relatively narrow feature map (the bottleneck); (2) uses a $1\times1$ convolution to expand the channels; (3) applies a depthwise $3\times3$ convolution to capture local spatial patterns; and (4) uses another $1\times1$ convolution to project back to a narrow bottleneck.

\begin{figure}
    \centering
    \vspace{-2mm}
    \includegraphics[width=\linewidth]{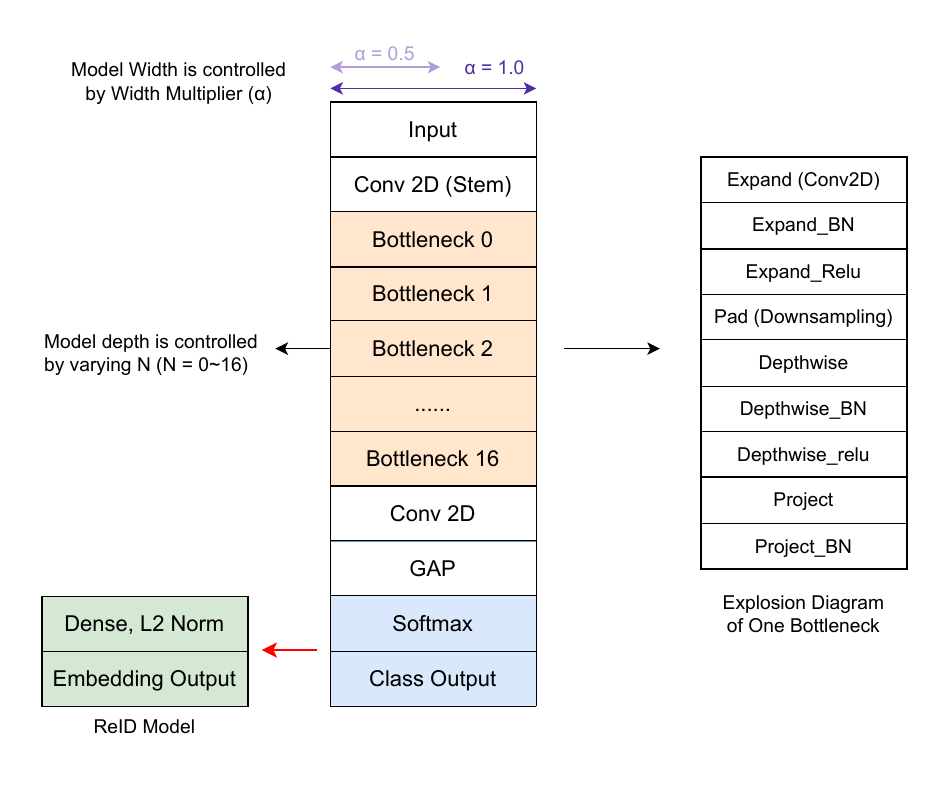}
    \vspace{-10mm}
    \caption{Architecture of the MobileNetV2-based Network with Controllable Depth and Width.}
    \vspace{-6mm}
    \label{fig:reid_model} 
\end{figure}

Overall, the MobileNetV2 backbone can be viewed as a sequence of $N=16$ such blocks arranged from shallow, high-resolution features to deeper, low-resolution features, as shown in \autoref{fig:reid_model}. In the standard MobileNetV2 design, a global width multiplier scales all channel counts in these blocks, and thus controls the overall model width. This regular ``stack of similar blocks'' layout is what we build on in Section~\ref{sec:depth-pruning}: we treat each inverted residual block as a unit that can be kept or removed when we adjust the model \textit{depth} (number of blocks), while the width multiplier controls the channel counts inside each block when we adjust the model \textit{width}.

Because Animal Re-ID is a retrieval task rather than a closed-set classification problem, we replace the final softmax classifier with a metric-learning head that produces a $d$-dimensional, $L_2$-normalised embedding, and we train the model with a triplet loss. This change is confined to the prediction layer and does not alter the backbone, so all width and depth scaling in the following subsections applies directly to the original MobileNetV2 design.

\begin{table*}[t]
\centering
\footnotesize
\renewcommand{\arraystretch}{1.1}
\setlength{\tabcolsep}{2pt}
\caption{Comparison of ReID performance of MobileNetV2 student models under different width multipliers $\alpha$.}
\vspace{-2mm}
\label{tab:different width multiplier results}
\begin{tabularx}{\textwidth}{lYYYYYYYYYYYY}
\toprule
& \multicolumn{4}{c}{ATRW} 
& \multicolumn{4}{c}{FriesianCattle2017} 
& \multicolumn{4}{c}{LionData} \\
\cmidrule(lr){2-5} \cmidrule(lr){6-9} \cmidrule(lr){10-13}
$\alpha$ 
& mAP & Top-1 & Top-5 & Top-10 
& mAP & Top-1 & Top-5 & Top-10 
& mAP & Top-1 & Top-5 & Top-10 \\
\midrule
0.25$^{1}$ 
& 10.1 & 10.4 & 36.8 & 54.7
& 18.5 & 22.4 & 49.4 & 67.1
& 11.0 & 11.5 & 32.8 & 54.5 \\
0.35$^{1}$ 
& 11.5 & 16.0 & 43.9 & 60.6
& 20.2 & 23.5 & 51.8 & 71.8
& 10.8 &  6.6 & 32.8 & 45.9 \\
0.35$^{2}$ 
& 30.7 & 66.7 & 86.1 & 91.5
& 58.9 & 72.9 & 94.1 & 96.5
& 13.3 & 14.8 & 37.7 & 59.0 \\
0.50$^{1}$ 
& 11.0 & 10.4 & 36.8 & 54.7
& 24.2 & 27.1 & 60.0 & 75.3
& 11.0 &  4.9 & 34.4 & 55.7 \\
0.50$^{2}$ 
& 28.8 & 62.0 & 82.5 & 87.5
& 59.2 & 76.5 & 92.9 & 97.6
& 14.2 & 14.8 & 37.7 & 57.4 \\
0.75$^{2}$ 
& 27.1 & 61.8 & 80.7 & 86.8
& 58.9 & 92.9 & 92.9 & 96.5
& 13.1 &  8.2 & 32.8 & 65.6 \\
1.00$^{2}$ 
& 24.3 & 55.3 & 77.6 & 85.4
& 57.2 & 69.4 & 87.1 & 90.6
& 13.0 & 11.5 & 36.1 & 49.2 \\
\midrule
& \multicolumn{4}{c}{MPDD} 
& \multicolumn{4}{c}{IPanda50} 
& \multicolumn{4}{c}{CoBRA ReID Youngstock} \\
\cmidrule(lr){2-5} \cmidrule(lr){6-9} \cmidrule(lr){10-13}
$\alpha$ 
& mAP & Top-1 & Top-5 & Top-10 
& mAP & Top-1 & Top-5 & Top-10 
& mAP & Top-1 & Top-5 & Top-10 \\
\midrule
0.25$^{1}$ 
&  3.6 &  1.9 & 12.5 & 20.2
& 15.5 & 16.8 & 53.7 & 71.1
&  9.8 & 12.4 & 40.4 & 59.4 \\
0.35$^{1}$ 
&  3.8 &  2.9 &  7.7 & 13.4
& 15.6 & 13.7 & 53.9 & 72.1
& 13.5 & 14.4 & 48.3 & 67.9 \\
0.35$^{2}$ 
& 30.1 & 40.4 & 59.6 & 78.8
& 18.8 & 39.2 & 73.5 & 87.0
& 42.6 & 67.7 & 90.4 & 95.2 \\
0.50$^{1}$ 
&  4.3 &  4.8 & 10.6 & 20.2
& 15.0 & 14.7 & 51.6 & 71.4
& 14.6 & 17.8 & 52.2 & 69.1 \\
0.50$^{2}$ 
& 29.5 & 39.4 & 68.3 & 76.0
& 20.0 & 38.1 & 74.2 & 84.7
& 45.0 & 63.1 & 88.6 & 94.1 \\
0.75$^{2}$ 
& 32.5 & 40.4 & 65.4 & 78.8
& 20.5 & 35.5 & 69.8 & 79.0
& 47.0 & 68.0 & 91.0 & 95.3 \\
1.00$^{2}$ 
& 31.5 & 37.5 & 66.3 & 79.8
& 21.3 & 36.7 & 73.0 & 83.7
& 45.7 & 66.5 & 91.4 & 95.3 \\
\bottomrule
\end{tabularx}

\vspace{0.5em}
\vspace{-3mm}
\begin{flushleft}
\footnotesize
$^{1}$ Trained without pre-trained weights.\,
$^{2}$ Trained with pre-trained weights.
\end{flushleft}
\vspace{-6mm}
\end{table*}

\vspace{-3mm}
\subsection{Reduce the Model Width}
\label{sec:width-pruning}

Our first question is how far we can narrow the MobileNetV2 backbone before accuracy breaks down. We keep the network topology fixed and adjust only the global width multiplier $\alpha$, which rescales each layer’s channels from $C_i$ to $\lfloor \alpha C_i \rfloor$ while leaving the block structure unchanged~\cite{howard2017mobilenets,sandler2018mobilenetv2,keras_mobilenet_api}. Therefore, $\alpha$ only changes the channel counts (the network width). The overall block structure remains the same while parameters and FLOPs are reduced~\cite{keras_mobilenet_api}.
Because our Animal Re-ID datasets are relatively small and inputs are only $64\times64$, we first ask how important ImageNet pre-training is compared with changing $\alpha$. TensorFlow provides ImageNet-pretrained checkpoints down to $\alpha = 0.35$, but not for narrower models. Rather than re-training every candidate on ImageNet (which would be costly), we adopt a realistic setup: architectures with available checkpoints use ImageNet pre-training, while those without are randomly initialised and trained only on the Animal Re-ID data. Concretely, we use pretrained models with $\alpha \in \{0.35, 0.5, 0.75, 1.0\}$ and train scratch models with $\alpha \in \{0.25, 0.35, 0.5\}$.

The results in Table~\ref{tab:different width multiplier results} show that pre-training has a much larger effect than width. At $\alpha = 0.35$, ImageNet initialisation raises mAP on ATRW from 11.5 (no pre-training) to 30.7, with gains of roughly 15–25 points across datasets compared to training from scratch. At $\alpha = 0.5$, the pretrained model again outperforms its scratch counterpart by about 18 mAP on ATRW, with similarly significant improvements observed across the other datasets. In contrast, the scratch model at $\alpha = 0.25$ reaches only 10.1 mAP on ATRW and similarly low scores on the other datasets, performing far below even the narrowest pretrained backbone. Thus, under limited Animal Re-ID data and $64\times64$ inputs, ImageNet pre-training is far more important for accuracy than moderate width changes, and very narrow models without pre-training are not competitive.

Having established that ImageNet pre-training is more important than moderate width changes, we now ask: among ImageNet-pretrained backbones, how sensitive is Re-ID performance to the choice of width multiplier $\alpha$ under our MCU memory constraints? Because the target platform is memory-limited, we restrict attention to the smaller ImageNet-pretrained MobileNetV2 variants with $\alpha \in \{0.35, 0.5, 0.75, 1.0\}$ and reuse the same metric-learning setup as before. These four models have sizes 6.18 MB, 7.31 MB, 9.89 MB, and 13.24 MB, respectively, so decreasing $\alpha$ yields a simple, monotonic trade-off between width and model size.

As shown in Table~\ref{tab:different width multiplier results}, our results show that once the backbone is ImageNet-pretrained, the narrow models have similar, and sometimes better performance than the wider ones. 
Varying $\alpha$ among pretrained models changes mAP by at most a few points, and the narrowest model ($\alpha = 0.35$) often matches or slightly exceeds the wider variants in many datasets. For example, on ATRW, the most challenging dataset, $\alpha = 0.35$ achieves the best mAP (30.7) and Top-1 (66.7), and on the remaining datasets its mAP is usually within a few points of the best configuration with comparable Top-1/Top-5 scores. This suggests that at $64\times64$ resolution the smallest pretrained MobileNetV2 already has sufficient capacity, so additional width brings limited benefit. Because width scaling shrinks MobileNetV2 from 13.24 MB ($\alpha = 1.0$) to 6.18 MB ($\alpha = 0.35$) with only minor losses in retrieval performance, we adopt the ImageNet-pretrained $\alpha = 0.35$ model as the student backbone for all subsequent experiments and deployment. Intuitively, low-resolution inputs and limited training data mean that the task does not require many distinct feature channels: a narrow pretrained backbone already captures most of the useful variation, so extra width mostly adds redundant capacity rather than improving discrimination.

\vspace{-2mm}
\subsection{Reducing the Model Depth}
\label{sec:depth-pruning}

Having fixed the width multiplier to $\alpha = 0.35$, the backbone still exceeds the memory budget of our embedded platform discussed in \autoref{sec:background}. We choose this sequential strategy because the preceding width study shows that the smallest ImageNet-pretrained width already provides a strong accuracy–size trade-off, while wider variants would further increase the memory pressure on the MCU. With this memory-first deployment objective, we fix the minimum pretrained width and focus the remaining search on depth reduction. We therefore next ask how much we can reduce the model depth, i.e. the number of inverted residual bottleneck blocks, while keeping the width multiplier unchanged. We will show that retaining only the first seven out of sixteen blocks yields a strong “knee point” in the accuracy–size trade-off: the model shrinks to 1.97 MB while remaining on the near-optimal mAP plateau, with performance comparable to the best depths on all datasets.


With the width fixed at $\alpha = 0.35$, depth is the other major degree of freedom in the backbone design. Increasing depth adds layers and therefore increases parameters and computation, but it is not clear how much additional depth actually helps Animal Re-ID at low resolution. We therefore investigate whether a shorter prefix of MobileNetV2 can retain most of the Animal Re-ID performance while substantially reducing model size.

To study this, we vary depth in a controlled way while keeping the rest of the architecture and training pipeline fixed. MobileNetV2 is organised as a sequence of inverted residual bottlenecks, so we view the $\alpha = 0.35$ backbone as a chain of 16 standard bottleneck blocks between an initial Stem and a final Head followed by the metric-learning embedding layers (global average pooling, dropout, fully connected, and $L_2$ normalisation). The Stem, the simplified bottleneck~0, the Head, and the embedding layers are kept fixed across all variants. These components define the backbone’s input resolution and embedding dimensionality. Keeping them unchanged preserves tensor shapes and maintains full compatibility with our loss functions and evaluation.

%
%
%

\begin{table*}
\centering
\footnotesize
\renewcommand{\arraystretch}{1.1}
\setlength{\tabcolsep}{2pt}
\caption{Performance comparison between Our Method and CLIP-ReID on six public datasets}
\vspace{-2mm}
\label{tab:final comparison results}
\begin{tabularx}{\textwidth}{lYYYYYYYYYYYY}
\toprule
& \multicolumn{4}{c}{ATRW} & \multicolumn{4}{c}{FriesianCattle2017} & \multicolumn{4}{c}{LionData} \\
\cmidrule(lr){2-5} \cmidrule(lr){6-9} \cmidrule(lr){10-13}
Method & mAP & Top-1 & Top-5 & Top-10 & mAP & Top-1 & Top-5 & Top-10 & mAP & Top-1 & Top-5 & Top-10 \\
\midrule
CLIP-ReID & 50.3 & 92.5 & 98.8 & 99.3
         & 73.4 & 88.2 & 100.0 & 100.0
         & 17.7 & 21.3 & 50.8 & 75.4 \\
Our Method & 47.7 & 87.3 & 95.5 & 97.2
           & 81.3 & 87.1 & 97.6 & 98.8
           & 23.8 & 37.7 & 59.0 & 67.2 \\
\midrule
& \multicolumn{4}{c}{MPDD} & \multicolumn{4}{c}{IPanda50} & \multicolumn{4}{c}{CoBRA ReID Youngstock} \\
\cmidrule(lr){2-5} \cmidrule(lr){6-9} \cmidrule(lr){10-13}
Method & mAP & Top-1 & Top-5 & Top-10 & mAP & Top-1 & Top-5 & Top-10 & mAP & Top-1 & Top-5 & Top-10 \\
\midrule
CLIP-ReID & 60.9 & 78.8 & 92.3 & 96.2
         & 27.5 & 77.8 & 94.3 & 96.7
         & 62.0 & 93.4 & 98.6 & 99.3 \\
Our Method & 59.4 & 73.1 & 90.4 & 93.3
           & 24.8 & 60.0 & 84.1 & 93.4
           & 50.6 & 74.7 & 94.6 & 97.3 \\
\bottomrule
\end{tabularx}
\vspace{-5mm}
\end{table*}

Starting from this baseline, we build a set of models that differ only in how many standard bottleneck blocks they contain. For each $N \in \{1,\ldots,16\}$, we keep the first $N$ standard bottlenecks, cut the network after block~$N$, and reuse the original Head and embedding layers on top of its output. As $N$ decreases, the backbone becomes shallower and the model size decrease, while the Stem, bottleneck~0, Head, and embedding layers remain unchanged. This construction isolates the effect of depth: by comparing these variants, we can see how retrieval accuracy degrades as we shorten the backbone and identify depths that offer a good accuracy–size trade-off for deployment.

\begin{figure}[t]
    \centering
    \includegraphics[width=\linewidth]{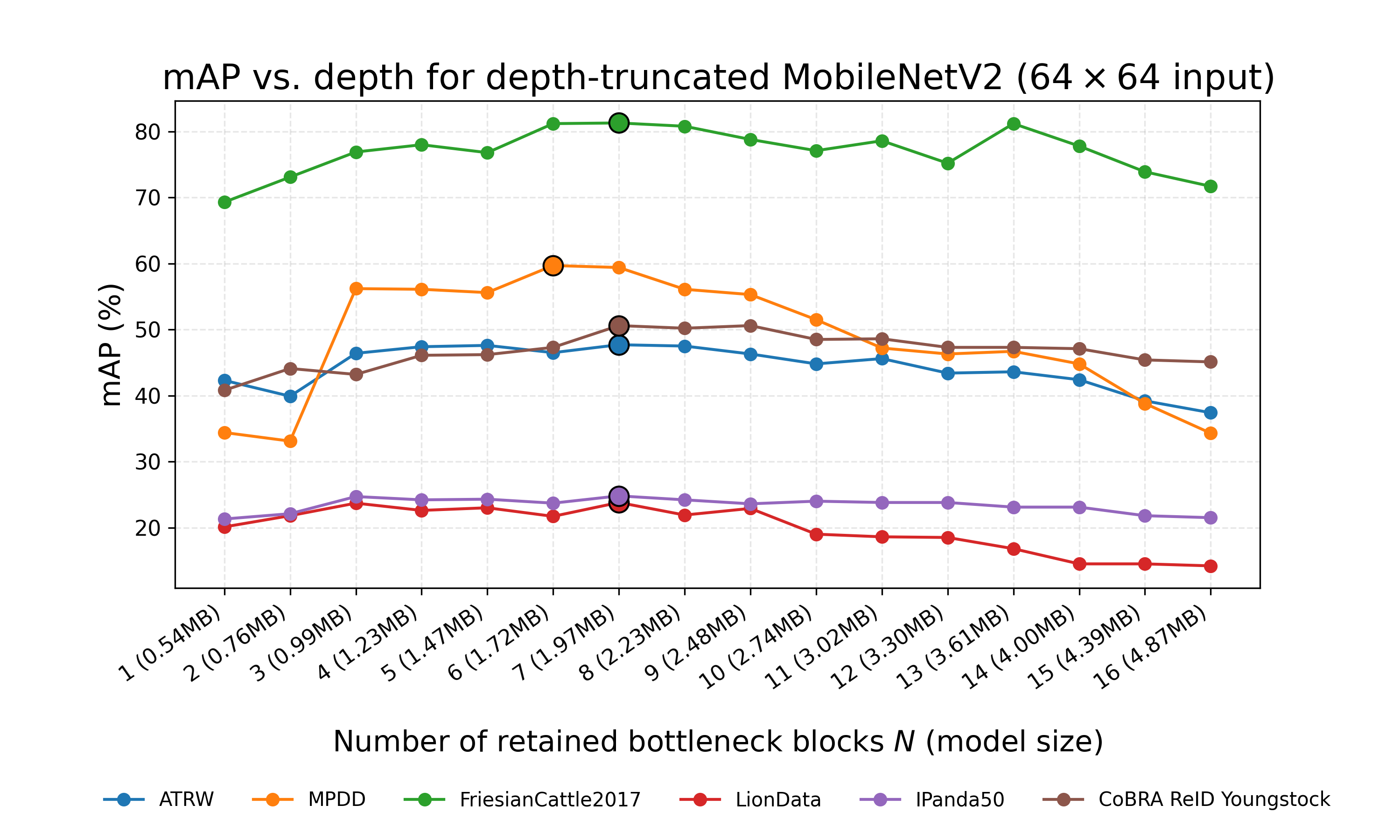}
    \caption{mAP with the number of retained bottleneck blocks $N$ for depth--truncated MobileNetV2 backbones.}
    \label{fig:depth_alpha35_plot}
    \vspace{-2mm}
\end{figure}

To evalute the different backbones, the width multiplier is fixed at $\alpha = 0.35$, and all inputs are resized to $64\times64$, matching the constraints of the target MCU deployment. For each value of $N$, we record mAP and model size.
\autoref{fig:depth_alpha35_plot} summarises the results. Across all datasets, a similar pattern emerges. For very shallow networks ($N{=}1$–3), increasing depth leads to large gains in mAP. Around moderate depths (approximately $N{=}4$–8), performance reaches a peak or plateau. Beyond this region, adding further blocks up to the full model ($N{=}16$) yields little or no improvement and can even slightly degrade mAP on some datasets. This indicates that, under low-resolution $64\times64$ inputs and narrow width, the representational capacity of the backbone saturates at moderate depth, and additional layers offer diminishing returns.

Among all depth configurations, $N{=}7$ emerges as a consistently strong choice. For ATRW, FriesianCattle2017, and CoBRA ReID Youngstock, $N{=}7$ is at or very near the global optimum, with mAP typically within 1–2 points of the best depth. On MPDD, the maximum mAP occurs at $N{=}6$, but the curve is flat and performance at $N{=}7$ is essentially unchanged. For LionData and IPanda50, where overall mAP is lower and the curves are flatter, $N{=}7$ also lies on the top plateau and is indistinguishable from the best configuration within normal experimental variation. Model size moves in the opposite direction: each additional block increases parameters and Flash usage. Within the $\alpha = 0.35$ family, the full-depth model ($N{=}16$) occupies about 4.87 MB, whereas $N{=}7$ uses only 1.97 MB, leading a reduction of nearly 60\% while staying on the accuracy plateau across datasets. We therefore regard $N{=}7$ as a deployment-friendly depth that offers a good balance between retrieval performance and compactness.

This knee point reflects the interaction between model capacity, input resolution, and data scale. Very shallow backbones underfit animal appearance variation, while moderate depth provides a larger receptive field and richer hierarchical features for retrieval.

%
However, with $64\times64$ inputs and fixed width $\alpha{=}0.35$, the repeated strided convolutions quickly reduce the spatial resolution of the feature maps. Beyond roughly seven bottleneck blocks, most additional layers operate on very small feature maps and mainly recombine existing coarse features rather than introducing genuinely new information, while still adding parameters. Given the limited amount of training data, this extra capacity is hard to exploit and can even hurt generalisation. The empirical curves in \autoref{fig:depth_alpha35_plot} therefore suggest that $N \approx 7$ strikes a good compromise: the backbone is deep enough to model the necessary appearance complexity, but not so deep that its capacity becomes redundant, making it well suited for MCU deployment.

\vspace{2mm}

\noindent\underline{Summary.} This section identifies the compact MobileNetV2 configuration used for deployment. The width study shows that ImageNet pre-training is more important than increasing channel width, making the smallest pretrained backbone, $\alpha=0.35$, the best accuracy--size trade-off. With this width fixed, the depth sweep identifies $N=7$ bottleneck blocks as a clear knee point, reducing Flash usage by almost 70

Compared with CLIP-ReID, the resulting model reduces size from 477,MB to 1.97,MB ($242\times$ smaller) while remaining competitive across the six public datasets (Table~\ref{tab:final comparison results}). These results support using the pruned MobileNetV2 backbone for embedded Animal Re-ID.



%% file: sections/experiment_on_arduino.tex
\vspace{-2mm}
\section{Experiment on Arduino}
\label{sec:sec5-experiment-arduino}

After pruning and compressing the backbone network as described in \autoref{sec:sec4-pruning}, we obtain a 1.97 MB FP32 model, which is then quantised for MCU deployment.. This section describes how this trained network is converted into an embedded-ready model through post-training quantisation, conversion into an embedded-ready model format, and deployment on a low-power Arduino platform for on-device evaluation.

\vspace{-2mm}
\subsection{Quantisation and Model Conversion}

The final deployment target is an Arduino-class MCU whose inference stack only supports efficient 8-bit integer (INT8) operations and does not execute standard 32-bit floating-point (FP32) operations commonly used in workstations and servers. To run our compact CNN under these constraints, we convert the pretrained FP32 model into a fully integer model and then export it into an embedded-ready format. At a high level, this involves (i) applying post-training quantisation to map weights and activations from FP32 to \texttt{INT8}, and (ii) converting the resulting integer-only model into a deployable MCU model.

We apply TensorFlow Lite Post-Training Quantization (PTQ) to convert the trained FP32 CNN into a fully \texttt{INT8} inference graph without additional retraining~\cite{tensorflow2022ptq, google2024ptq}. The converter uses integer-only optimisation, with per-channel quantisation for convolutional weights and per-tensor quantisation for activations~\cite{nagel2021white}. A representative set of 100 randomly sampled training images, preprocessed as during training, is used to calibrate activation ranges and determine tensor scales and zero-points.

The resulting quantised network is exported as a single \texttt{.tflite} FlatBuffer and executed directly by TensorFlow Lite Micro on the Arduino. This reduces the model size from 1.97 MB (FP32) to approximately 84 KB (INT8). Since larger variants did not provide consistent accuracy gains in Section~\ref{sec:sec4-pruning}, we use this compact INT8 model for all deployments.

\vspace{-2mm}
\subsection{Deployment}

After obtaining the quantised model, we deploy it to an Arduino Nano 33 BLE Sense MCU to demonstrate fully on-device inference. The quantized model and a test image are compiled into the firmware as constant arrays, so that all computation runs locally on the device without any external host. The input image is resized to \(64 \times 64 \times 3\) and quantised to \texttt{int8} using the same preprocessing and scaling procedure as in training, ensuring consistency between the training and deployment pipelines.
At runtime, the model is executed using TensorFlow Lite for MCUs, which provides a lightweight interpreter and a small pre-allocated memory arena in SRAM for all tensors. During inference, the pre-quantized image array is copied into the model’s input buffer and a single forward pass is performed, producing an \texttt{int8} feature embedding at the network output.

For analysis and retrieval, this embedding is converted back to a continuous representation using the standard quantitation rule, 
$\text{real} = (\text{q} - \text{zero\_point}) \times \text{scale},$
applied element-wise with the scale and zero-point parameters associated with the output tensor. The resulting floating-point feature vector serves as a compact representation of the input image and is used for similarity measurement and recognition in downstream evaluation. Empirically, this deployment pipeline performs stable end-to-end inference on the resource-constrained microcontroller, confirming the practicality of the proposed TinyML model in real embedded environments.

\vspace{-2mm}
\subsection{Data Collection and Evaluation}

Our goal in this experiment is to evaluate whether the proposed TinyML model can perform reliable Animal Re-ID when deployed end-to-end on a real MCU. To approximate a realistic deployment scenario, we evaluate the model on a self-collected cattle dataset recorded in the natural grazing environment of in Auckland, New Zealand.
In this local data collection effort, we obtained 45 individual cattle identities and 549 images in total. After preprocessing, quality filtering, and manual annotation, the images were split into a training set and an evaluation set using an approximate 80\%/20\% ratio. The final dataset contains 36 training identities and 9 evaluation identities, corresponding to 448 training images, 65 gallery images, and 36 query images. 

For evaluation, we use the same gallery–query protocol described earlier: identities are split into a gallery set (reference images) and a query set (images to be identified). Gallery images are passed through the on-device model to extract embeddings, which are stored on the Arduino as a reference database. Each query image is then processed to obtain its embedding and matched to the gallery via nearest-neighbour retrieval on the device. To assess the impact of quantisation and deployment, we compare two settings on the same splits: \textit{Cluster}, the original FP32 model run on a compute cluster as an upper bound, and \textit{Arduino}, the quantised INT8 model running entirely on the Nano 33 BLE Sense, with both inference and matching on the MCU.

\begin{table}[t]
    \centering
    \footnotesize
    \caption{Server and Arduino performance.}
    \vspace{-2mm}
    \label{tab:cluster_arduino}
    \resizebox{\columnwidth}{!}{
        \begin{tabular}{lccccc}
        \toprule
        \textbf{Method} & \textbf{mAP} & \textbf{Top-1} & \textbf{Top-5} & \textbf{Top-10} & \textbf{Model Size} \\
        \midrule
        Cluster & 50.1 & 61.1 & 86.1 & 94.4 & 1.97 MB \\
        Arduino & 45.6 & 61.1 & 96.1 & 97.2 & 84 KB \\
        \bottomrule
        \end{tabular}
    }
    \vspace{-5mm}
\end{table}

The results in ~\autoref{tab:cluster_arduino} show how much recognition performance is preserved when moving from the floating-point cluster model to the heavily compressed, integer-only MCU model.
The cluster model provides the reference performance in FP32. When deployed on the Arduino, the quantised model is compressed from {1.97 MB} to only {84 KB}, yet it is able to retain most of the Animal Re-ID capability. The mAP decreases moderately (50.1 to 45.6), indicating a small loss in ranking quality, which is consistent with the small numerical changes introduced when compressing the model to 8-bit integers. However, the Top-1 accuracy remains identical at 61.1, showing that the top prediction is correct at the same rate in both settings, and the Top-5 and Top-10 accuracies fluctuate only slightly while staying within a high-performance range.
Since prior Animal Re-ID studies have mainly evaluated server- or GPU-side inference, there is no established MCU-specific accuracy target for this setting. Therefore, our deployment goal is to preserve the cluster-side retrieval performance as much as possible under the Arduino memory and integer-only inference constraints.
These results indicate that the proposed quantisation and deployment pipeline can reduce the model size by more than an order of magnitude while preserving comparable Animal Re-ID performance.

\vspace{-2mm}
\subsection{Fine Tuning}

We have seen that training a model from scratch on our self-collected cattle dataset can yield good performance. However, this setup implicitly assumes that it is always possible to gather a large number of images and train a new model for every deployment site. In practice, this is rarely feasible: gathering and annotating hundreds of images per animal is time-consuming, requires expert effort, and demands compute resources that may not be available in the field.
At the same time, there are several public Animal Re-ID datasets. For example, the FriesianCattle2017 dataset also contains cattle. This naturally raises the question of whether we can reuse a model trained on such public data and adapt it to a new site with only a small amount of local data.

\begin{figure}[t]
    \centering
    \subfloat[Friesiancattle2017\label{fig:public_cattle}]{%
        \includegraphics[width=0.36\columnwidth,height=3cm]{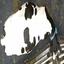}}
    \hspace{8mm}
    \subfloat[Collected Cattle Image\label{fig:new_cattle}]{%
        \includegraphics[width=0.36\columnwidth,height=3cm]{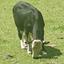}}
    \caption{Cattle from different regions.}
    \vspace{-6mm}
    \label{fig:comparison_cattle}
\end{figure}

However, training on a public dataset and then deploying the model as-is at a new site often does not work well. Public datasets and real deployment environments can differ substantially. As illustrated in Figure~\ref{fig:comparison_cattle}, cattle from different regions are photographed under different lighting and weather conditions, with distinct backgrounds, ground surfaces, and camera viewpoints. In addition, the animals themselves may look different due to age, body condition, feeding practices, or surface contamination such as mud or partial occlusion~\cite{beery2018recognition, rastikerdar2024situ, burkard2024automated, tuia2022perspectives}. These differences create a domain shift between the source dataset and the target deployment site~\cite{zhou2022domain}. If we train only on the public dataset and apply the model directly to our collected images, performance drops because the model has never seen the visual characteristics of the new environment. This is shown in \autoref{tab:finetune_results}.

\begin{table}
    \centering
    \caption{Performance under different fine-tuning}
    \vspace{-1mm}
    \label{tab:finetune_results}
    \footnotesize
    \begin{tabular}{lcccc}
        \hline
        \textbf{Method} & \textbf{mAP} & \textbf{Top-1} & \textbf{Top-5} & \textbf{Top-10} \\
        \hline
        No Fine-tuning                     & 45.0 & 72.2 & 88.9 & 94.4 \\
        Fine-tuned (few-shot)              & 47.8 & 75.0 & 88.9 & 97.2 \\
        Training from scratch (new data)   & 50.1 & 61.1 & 86.1 & 94.4 \\
        \hline
    \end{tabular}
    \vspace{-6mm}
\end{table}

To bridge this gap without collecting a large new dataset, we adopt a data-efficient fine-tuning strategy. We first train the model on a public dataset to learn a general animal re-identification representation and save the resulting weights. When adapting to our collected data, we reload this pretrained model, freeze the backbone network, and fine-tune only the final embedding layer using the target images. During this stage, we use only a very small amount of new data, typically two to three images per animal identity. Since the target-site data are extremely limited, updating even part of the backbone may overfit to the few available examples and weaken the general representation learned from the public dataset. We therefore restrict adaptation to the final embedding layer, which provides a lightweight way to adjust the feature space to the new site while keeping the backbone stable. This preserves the generic features learned from the source domain, allows the last layer to adjust to the new site, and greatly reduces annotation and computation compared with training from scratch.

Table~\ref{tab:finetune_results} compares three strategies on the collected dataset. The No fine-tuning setting applies the model trained on the public dataset directly to Cornwall Park, with no adaptation. The Fine-tuned (few-shot) setting uses the same pretrained model but fine-tunes only the embedding layer with three images per identity. The Training from scratch (new data) setting corresponds to the model trained solely on the collected data, without using the public dataset. The results show that few-shot fine-tuning consistently improves performance over the no–fine-tuning baseline: both mAP and Top-1 increase when a small number of collected data images is used for adaptation. At the same time, the fine-tuned model remains competitive with the model trained from scratch, despite using far fewer local images and much less computation. In particular, the few-shot model attains the highest Top-1 accuracy, while its mAP stays close to the from-scratch upper bound. Overall, this indicates that the proposed fine-tuning strategy provides a practical, data-efficient way to adapt animal re-identification models to new deployment sites without full retraining.

%% file: sections/discussion.tex
\vspace{-3mm}
\section{Discussion}


This paper presents an initial step toward Animal Re-ID on MCUs, but there are several limitations we'd like to discuss here. 

\textit{First}, our treatment of low-resolution inputs is somewhat idealised. Most experiments use images downsampled to $64 \times 64 \times 3$ from public datasets or locally collected smartphone photos, while Arduino-class camera modules may introduce additional distortions such as colour shifts, limited dynamic range, motion blur, and compression artefacts that are not explicitly modelled. Our focus is to examine whether an MCU-scale model can handle low-resolution inputs under tight memory budgets. In addition, the cattle deployment dataset is relatively small, with limited evaluation identities and query images, so the reported retrieval scores should be viewed as an initial feasibility result rather than a statistically exhaustive deployment benchmark. Larger-scale evaluations across more animals, farms, capture conditions, and embedded camera pipelines are needed to better quantify robustness in practical deployments.

\textit{Second}, at the system level, our current system is not yet a fully automated end-to-end pipeline. Deploying the model still requires several manual steps before exporting the quantised model to the microcontroller, and we do not target a single plug-and-play network that works out of the box across all farms or environments. Some site-specific data collection and light tuning are still expected. Future work can broaden on-device experiments under realistic embedded imaging conditions and integrate the method into a more automated data handling, adaptation, and deployment pipeline.

%% file: sections/related_work.tex
\vspace{-2mm}
\section{Related Work}

\noindent\textbf{Knowledge distillation and compact Re-ID.}
Knowledge distillation (KD) is widely used for model compression and has been applied to lightweight Re-ID by transferring similarity structures or uncertainty-aware features from stronger teachers~\cite{hinton2015distilling,wu2019distilled,jin2020uncertainty}. KD, quantization, and pruning are also common tools for TinyML deployment~\cite{heydari2025tiny}. Unlike these settings, our target combines fine-grained Animal Re-ID, low-resolution sensing, and sub-MB MCU memory limits. We therefore evaluate several KD routes and find that conventional feature-level KD brings limited gains over directly trained compact CNNs, motivating direct structure scaling and quantization.

\noindent\textbf{Efficient architectures for TinyML.}
Compact architectures and hardware-aware design methods improve accuracy--efficiency trade-offs for edge vision. MobileNetV2 and EfficientNet reduce computation through efficient convolutional design and model scaling~\cite{sandler2018mobilenetv2,tan2019efficientnet}, while MCUNet, Once-for-All, and FBNet further search or co-design models for device-specific memory and latency constraints~\cite{lin2020mcunet,cai2019once,wu2019fbnet}. These methods mainly target classification or general efficient vision, whereas our work focuses on retrieval-based Animal Re-ID on MCU-class hardware, where the model must produce discriminative embeddings from degraded $64\times64$ inputs within a strict microcontroller memory budget.

%% file: sections/conclusion.tex
\vspace{-3mm}
\section{Conclusion}

We take a first step toward running Animal Re-ID directly on MCU-class devices by studying the full path from compression diagnosis to model design and on-device retrieval. Starting from the gap between GPU-oriented models and MCU constraints, we explore architecture scaling and quantisation under low-resolution sensing and strict Flash/SRAM budgets, arriving at a tiny INT8 CNN that fits within MCU memory limits while maintaining competitive accuracy on both public benchmarks and a real-world cattle deployment. Building on this backbone, we show that a simple few-shot fine-tuning scheme can adapt the model to new sites using only a small number of images per animal, avoiding the need to train from scratch at each location. Together, these results outline a practical recipe for MCU-based Animal Re-ID and point toward future systems that perform monitoring entirely on low-power edge hardware.